\documentclass[conference]{IEEEtran}

\usepackage{cite}
\usepackage{amsmath,amssymb,amsfonts}
\usepackage{algorithmic}
\usepackage{graphicx}
\usepackage{textcomp}
\usepackage{xcolor}
\def\BibTeX{{\rm B\kern-.05em{\sc i\kern-.025em b}\kern-.08em
    T\kern-.1667em\lower.7ex\hbox{E}\kern-.125emX}}
\begin{document}

\title{A Survey on Knowledge Graph Structure and Knowledge Graph Embeddings}

\author{\IEEEauthorblockN{1\textsuperscript{st} Jeffrey Sardina}
\IEEEauthorblockA{\textit{School of Computer Science \& Statistics} \\
\textit{Trinity College Dublin}\\
Dublin, Ireland \\
0000-0003-0654-2938}
\and
\IEEEauthorblockN{2\textsuperscript{nd} John D. Kelleher}
\IEEEauthorblockA{\textit{School of Computer Science \& Statistics} \\
\textit{Trinity College Dublin}\\
Dublin, Ireland \\
0000-0001-6462-3248}
\and
\IEEEauthorblockN{3\textsuperscript{rd} Declan O'Sullivan}
\IEEEauthorblockA{\textit{School of Computer Science \& Statistics} \\
\textit{Trinity College Dublin}\\
Dublin, Ireland \\
0000-0003-1090-3548}
}

\maketitle

\begin{abstract}
Knowledge Graphs (KGs) and their machine learning counterpart, Knowledge Graph Embedding Models (KGEMs), have seen ever-increasing use in a wide variety of academic and applied settings. In particular, KGEMs are typically applied to KGs to solve the link prediction task; i.e. to predict new facts in the domain of a KG based on existing, observed facts. While this approach has been shown substantial power in many end-use cases, it remains incompletely characterised in terms of how KGEMs react differently to KG structure. This is of particular concern in light of recent studies showing that KG structure can be a significant source of bias as well as partially determinant of overall KGEM performance. This paper seeks to address this gap in the state-of-the-art. This paper provides, to the authors' knowledge, the first comprehensive survey exploring established relationships of Knowledge Graph Embedding Models and Graph structure in the literature. It is the hope of the authors that this work will inspire further studies in this area, and contribute to a more holistic understanding of KGs, KGEMs, and the link prediction task.
\end{abstract}


\section{Introduction} \label{sec-kgs-hyps-and-lp}

\begin{figure*}
  \centering
  \includegraphics[width=0.9\textwidth]{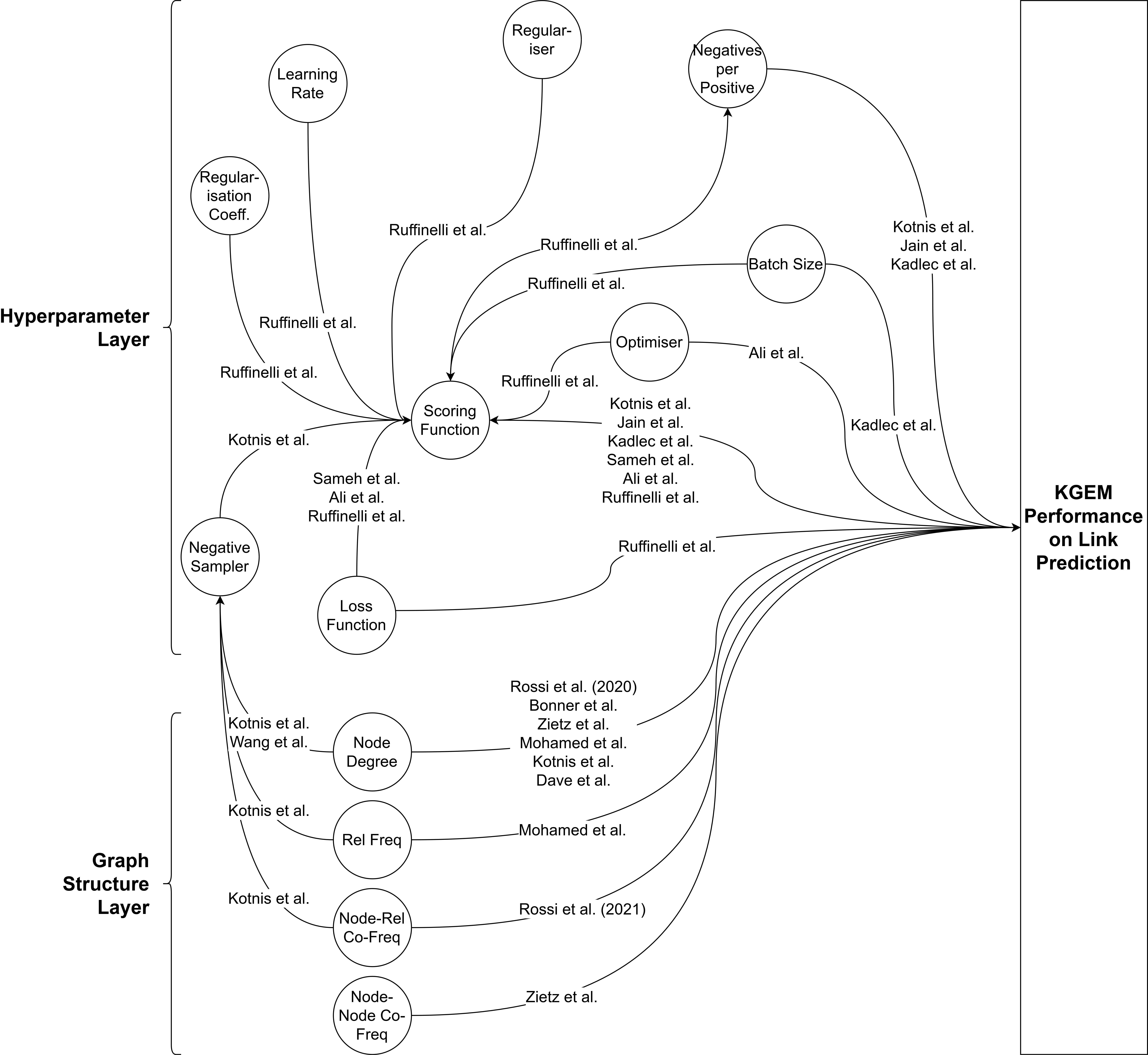}
  \caption{An overview of hyperparameter and graph structure influence on KGEM performance on the link prediction task, annotated by evidence in the literature. A directed edge from a subject node to an object node indicates that the item subject is dependent upon the object. Nodes are connected to Link Prediction Performance if they have been shown to ave concrete influence on overall link prediction performance. Note that as the literature has shown that hyperparameters are generally dependent on both the KGEM and KG used \cite{dark-into-light,old-dog-new-tricks}, only hyperparameters that have been \textit{specifically} studied are included.}
  \label{fig-hyp-dependencies}
\end{figure*}

Knowledge Graphs (KG) are databases that represent data in a graphical format \cite{kgs-overview}. In KGs, all data can be represented as statements called triples in the form $(s,p,o)$ consisting of a subject (or ``head") node $s$ an object (or ``tail") node $o$, and a directed predicate (or ``relationship") edge $p$ describing how the subject and object relate \cite{kgs-overview}.

Knowledge Graph Embedding Models (KGEMs) are machine learning models that learn to represent the information content of a KG in vector space. These vector embeddings are then used to make predictions about, and reason on, data in the graph \cite{rml-review,kge-survey}. In particular, KGEMs typically are used to solve what is called the Link Prediction (LP) task. In LP, the goal is to predict the object (or subject) of a triple given the subject (or object) and the relationship. For example, a KGEM could attempt to predict the link prediction query $(Frodo, is-from, ?)$ if it was trained on a KG describing \textit{The Lord of the Rings}.

While many existing studies have shown that KGEM performance is influenced by, and often biased by, elements of KG structure \cite{do-kges-learn-rels,embedding-position-centrality,topological-imbalance,edge-prob-due-to-node-deg,popularity-agnostic-eval,neg-sampler-analysis,kgs-overview,old-dog-new-tricks,centrality-measures,kges-for-lp-compare}, no study known to the authors has attempted to aggregate all findings relating to KG structure and KGEM performance to date. Finally, while various hyperparameter choices (including model component choices such as the choice of negative sampler or loss function) in a KGEM are known to interact with elements of structure, no study known to the authors has attempted to provide a general survey of such structure-based hyperparameter preference in KGEM literature.

This paper seeks to address both of these gaps in the state-of-the-art by giving an overview of state-of-the-art studies that have documented relationships between KG structure, KGEM hyperparameters, and KGEM performance on the link prediction task. It begins in Section \ref{sec-kg-struct-measurse} with a description of the most common measures of KG structure in the literature. Following this, it describes in Section \ref{sec-hyp-performance-analysis} how these features relate to KGEM performance and KGEM hyperparameter preference. It concludes with a list of open questions to encourage future work in this area. A summary of the main results of this survey is shown in Figure \ref{fig-hyp-dependencies}.

We note that a description of KGEMs and their components / hyperparameters is beyond the scope of this article. For such information, the reader is directed to Nickel et al. \cite{rml-review} and Wang et al. \cite{kge-survey}. Discussion of effects on \textit{ontological structure} of a KG is also out fo the scope of this work, but can be found in Ali et al. \cite{dark-into-light}. In terms of notation, we use the terms ``KGEM" and ``scoring function" interchangeably to refer to the model being used (such as TransE, DistMult, or ComplEx).

\section{Measures of Knowledge Graph Structure} \label{sec-kg-struct-measurse}
This section surveys how KG structure, and its impact on link prediction, is characterised in the literature. Table \ref{tab-struct-metrics} gives an overview of the structural metrics discussed in the literature and the manner in which they were referenced, ranging from * (mentioned but not analyses), ** (used to characterise KGs), and *** (used to characterise KGs and LP).

\begin{table*}[!ht]
    \centering
    \begin{tabular}{|l|p{1.4cm}|p{1cm}|p{1.7cm}|p{1.7cm}|p{1.4cm}|p{1.3cm}|}
    \hline
        \textbf{Article} & \textbf{Node Degree} & \textbf{Rel Freq} & \textbf{Node-Rel Co-Freq} & \textbf{Node-Node Co-Freq} & \textbf{Other (Node)} & \textbf{Other (Rel)} \\ \hline
        Rossi et al. (2020) \cite{do-kges-learn-rels} & *** & ** & . & . & . & . \\ \hline
        Sadeghi et al. (2021) \cite{embedding-position-centrality} & ** & . & . & . & *** & . \\ \hline
        Bonner et al. (2022) \cite{topological-imbalance} & *** & ** & . & . & . & . \\ \hline
        Zietz et al. (2024) \cite{edge-prob-due-to-node-deg} & *** & . & ** & *** & *** & *** \\ \hline
        Mohamed et al. (2020) \cite{popularity-agnostic-eval} & *** & *** & ** & ** & . & . \\ \hline
        Kotnis et al. (2017) \cite{neg-sampler-analysis} & *** & *** & *** & . & . & . \\ \hline
        Hogan et al. (2021) \cite{kgs-overview} & ** & . & . & . & ** & ** \\ \hline
        Ruffinelli et al. (2020) \cite{old-dog-new-tricks} & * & * & . & . & . & . \\ \hline
        Dörpinghaus et al. (2022) \cite{centrality-measures} & ** & . & . & . & ** & . \\ \hline
        Rossi et al. (2021) \cite{kges-for-lp-compare} & ** & ** & *** & . & . & *** \\ \hline
        Sardina et al. (2022) \cite{centrality-and-hyperparams} & *** & . & . & . & *** & . \\ \hline
        Sardina et al. (2024) \cite{twig} & *** & *** & *** & *** & . & . \\ \hline
        Dave et al. (2024) \cite{struct-impact-on-kges} & *** & . & . & . & . & . \\ \hline
    \end{tabular}
    \caption{An overview of papers describing KG structure in the state-of-the-art, what structural metrics they used, and how they used them. Key: *** = used to characterise both KGs and link prediction on those KGs; ** = used to characterise KGs; * = mentioned but not discussed; . =  not mentioned. Abbreviations: Rel = relation; Freq = frequency.}
    \label{tab-struct-metrics}
\end{table*}

Most literature that describes Knowledge Graph structure focuses on frequency-based metrics. Specifically, the four core metrics that appear consistently in KG / link prediction literature are:

\begin{itemize}
    \item \textbf{Degree}: The degree of a node is the number of relationships that connect to it. In other words, it is the number of times that node appears as either a subject or an object in a triple in the KG.
    \item \textbf{Relationship Frequency}: The frequency of a relationship is the number of times it is present in a triple in a KG.
    \item \textbf{Node-Relationship Co-Frequency}: Node-relationship co-frequency is the number of times that a given node and relationship co-occur in the same triples in a KG.
    \item \textbf{Node-Node Co-Frequency}: Node-node co-frequency is the number of times that two nodes co-occur in the same triples in a KG.
\end{itemize}

All of these metrics have been discussed in link prediction literature as relevant to the link prediction task in at least one publication, and noted as relevant to characterising KG structure in at least two publications, as outlined in Table \ref{tab-struct-metrics}.

The majority of existing publications in the area of KG structure show that node-based characterisation of KGs is more common than relationship-based characterisation in the literature, and is taken as generally representative of KG connectivity \cite{do-kges-learn-rels,embedding-position-centrality,topological-imbalance,edge-prob-due-to-node-deg,popularity-agnostic-eval,neg-sampler-analysis,kgs-overview,old-dog-new-tricks,centrality-measures,kges-for-lp-compare}. Relationship-based metrics, while highly relevant, tend to be considered less. Similarly, Co-frequency based metrics, while shown to be relevant in the literature \cite{edge-prob-due-to-node-deg,popularity-agnostic-eval,neg-sampler-analysis}, are less-heavily used but remain generally relevant.

Many other structural features have been considered in the literature. Dörpinghaus et al. (2022) \cite{centrality-measures}, Hogan et al. (2021) \cite{kgs-overview}, and Zietz et al. (2024) \cite{edge-prob-due-to-node-deg} are particularly notable in giving a detailed descriptions of more complex measures of KG structure. However, they tend to focus on \textit{different} metrics, resulting in little consensus in the literature for use of these other metrics for characterising KGs and LP performance.

Of all the studies outlined in Table \ref{tab-struct-metrics}, we focus on those that provide the most insight into the commonly-used frequency-based metrics outlined above. These are: Rossi et al. (2020) \cite{do-kges-learn-rels}, Sadeghi et al. (2021) \cite{embedding-position-centrality}, Bonner et al. (2022) \cite{topological-imbalance}, Zietz et al. (2024) \cite{edge-prob-due-to-node-deg}, Mohamed et al. (2020) \cite{popularity-agnostic-eval}, Kotnis et al. (2017) \cite{neg-sampler-analysis}, and and Rossi et al. (2021) \cite{kges-for-lp-compare}. A summary of each of these studies follows.

\subsubsection{Rossi et al. (2020)}
Rossi et al. (2020)  performs a KG-structure-based evaluation of the KGEMs TransE and DistMult on FB15k, FB15k-237, WN18, and WN18RR \cite{do-kges-learn-rels}. Specifically, they examine whether the degree of a node influences how well that node is learned and can be used in link prediction \cite{do-kges-learn-rels}. Their results indicate that, on both TransE and DistMult on all datasets tested, higher degree nodes are learned substantially better than lower-degree nodes \cite{do-kges-learn-rels}. They further highlight that existing KGEM benchmark datasets (FB15k, FB15k-237, WN18, and WN18RR) can exaggerate KGEM performance because high-degree entities are over-represented in their test sets, which means that KGEMs that learn to predict only a few nodes well can still appear to have high performance \cite{do-kges-learn-rels}.

While their analysis is primarily focused on node degree and its resulting impacts on link prediction, they use both node degree and relationship frequency to profile KG structure \cite{do-kges-learn-rels}. Importantly, they show that these common KG benchmark datasets exhibit extreme skew in node degree and relationship frequency in both the training and the test sets, something they show biases link prediction evaluation. They further highlight that low-degree nodes and low-frequency relations have less information about them that can be learned by link predictors, an effect they characterise both through re-analysis of KGEM evaluation and through an analysis of embedding space \cite{do-kges-learn-rels}.

Specifically, through an analysis of the position of nodes in embedding space, they highlight that higher-degree nodes tend to be more isolated from other nodes because there is more information about them in the graph to allow for high-quality representation and distinction from other nodes in embedding space \cite{do-kges-learn-rels}. They further note that lower-degree nodes tend to be very close to each other in embedding space, meaning that they cannot be as readily distinguished from each other \cite{do-kges-learn-rels}. The result of this is that link prediction queries asked to predict a high-degree node will be more successful than those that are asked to predict a low-degree node \cite{do-kges-learn-rels}.

The result of their analysis is a very clear conclusion that node degree is, at least in part, a determinant of link prediction performance and thereby highly relevant to characterising KG structure and link prediction \cite{do-kges-learn-rels}.

\subsubsection{Sadeghi et al. (2021)}
Sadeghi et al. (2021) build a GNN-based link predictor called GFA-NN that explicitly models node centrality and relative node position\cite{embedding-position-centrality}. They model node centrality specifically using Katz Centrality, which is a generalised version of node degree that accounts for the degrees of nodes nearby every node as well \cite{embedding-position-centrality}. They show that their model can match or exceed the performance of KGEMs on link prediction, and attribute this increased performance to its ability to model KG structure.

While the Sadeghi et al. (2021) paper is focused mostly on GNN-based link prediction, it is included in this analysis because of how they annotate the KGEM baselines they use. They suggest that GFA-NN is able to beat KGEM baselines on WN18RR but not on FB15k-237 because WN18RR has a wider distribution of node centrality values \cite{embedding-position-centrality}. They suggest that on FB15k-237, where degrees are more consistent and less spread out, that KGEMs are better able to learn -- thus providing some evidence that node centrality is key to understanding how KGEMs learn \cite{embedding-position-centrality}.

Finally, they call out KGEMs specifically for learning based on (only) a ``1-hop neighbourhood" around each node -- suggesting that they are only able to learn from very localised graph features \cite{embedding-position-centrality}. Considering their findings on FB15k-237 and WN18RR \cite{embedding-position-centrality}, as well as Rossi et al. (2020)'s findings that FB15k-237 and WN18RR have a massive skew in degree values  \cite{do-kges-learn-rels}, this is consistent with the idea that KGEMs learn better in more dense, connected regions of a graph.

\subsubsection{Bonner et al. (2022)}
Bonner et al. (2022) takes a similar approach to Rossi et al. (2020) \cite{do-kges-learn-rels} in that they both identify significant skew of node degrees in KG, and examine how this can lead to degree-related biases in link prediction using KGEMs \cite{topological-imbalance}. They perform their analysis specifically in the biomedical context, and show that degree imbalance, and degree-biased in link prediction, remains very common and very problematic in that domain.

Specifically, Bonner et al. (2022) calls out node degree as the most commonly used topological measure (although they do cite other works mentioned that relationship frequency is relevant as well) \cite{topological-imbalance}. They extracted information from the biomedical KG HetioNet and learned the graph using the KGEMs TransE, TransH, ComplEx, RotatE and DistMult \cite{topological-imbalance}.

They then examined the ranked list for predicting which genes were associated with given disease \cite{topological-imbalance}. Taking the ranked list outputs of these predictions, they find that higher-degree gene nodes tend to be preferentially predicted across all 137 diseases tested \cite{topological-imbalance}. They additionally found that high-degree nodes that were not directly observed to be connected to the disease of interest in the KG would be preferred over lower-degree nodes that had been observed in the KG training set to be linked to the disease of interest \cite{topological-imbalance}. This last point particularly is of note, suggesting that KGEMs overfit based on degree even to the point of disregarding other observed connections in the dataset \cite{topological-imbalance}.

Finally, they show that deleting edges incident on a node (so as to lower its degree) result in it being considered less plausible as an answer in link prediction and that adding edges similarly could make a node be considered more plausible as an answer \cite{topological-imbalance}. Taken together, this suggests that node degree is largely influential on link prediction outputs in KGEMs.

\subsubsection{Zietz et al. (2024)}
Zietz et al. (2024), like Rossi et al. (2020) \cite{do-kges-learn-rels}  and Bonner et al. (2022) \cite{topological-imbalance} highlights that KGs tend to have very skewed degree distributions and performs an analysis of degree-biased bias in KGs \cite{edge-prob-due-to-node-deg}. They do this by asking if degree (as well as some other structural methods) are sufficient to allow for link prediction on their own \cite{edge-prob-due-to-node-deg}. The system they create uses one of several structural features -- typically based on degree or node-node co-frequency -- to estimate the probability that an edge should exist \cite{edge-prob-due-to-node-deg}.

To do this, they take the KG HetioNet and split it into distinct sub-graphs in which only one edge type is present (meaning that each individual sub-graph is effectively an unlabelled graph, not a true KG) \cite{edge-prob-due-to-node-deg}. They then take each of these unlabelled networks and construct what they call an ``edge prior" that calculates edge probability as a function of various frequency-based properties of the graph \cite{edge-prob-due-to-node-deg}. Their evaluation shows that they are able to reconstruct each of these unlabelled graphs with very high accuracy using this method, suggesting that simple graph structural features are sufficient to predict links in unlabelled graphs \cite{edge-prob-due-to-node-deg}.

It is important to reiterate that, while the unlabelled graphs they used are extracted from a Knowledge Graph, they are not multi-relational KGs in the form present in standard KG literature \cite{edge-prob-due-to-node-deg}. As such, these results must be interpreted with care in the context of KGs and KGEMs in the general case.

\subsubsection{Mohamed et al. (2020)}
Mohamed et al. (2020) establish that both node degree and edge frequency are subject of heavy skew in Knowledge Graphs, with there being many nodes / edges with low frequency, and very few that have very high frequencies \cite{popularity-agnostic-eval}. They are particularly notable for showing that this skew follows a power law, which they mathematically annotate and describe in the context of the benchmark KGs FB15k, WN18, and YAGO3-10 \cite{popularity-agnostic-eval}.

The bulk of the paper then focuses on how to re-define evaluation metrics to assign lower-weight to higher frequency nodes / relations as a method of re-balancing test set to stratify evaluation equally across all nodes and relations \cite{popularity-agnostic-eval}. They first provide evidence that the degree of subject, predicate, and object items in a triple are not correlated -- i.e. the presence of a high-degree subject does not imply the presence of a high (or low) frequency predicate, nor the presence of a high (or low) degree object \cite{popularity-agnostic-eval}. Because of this, they note that re-weighting of the test set cannot be done directly at the triple level -- there is no way to label a triple as "over-represented" or ``under-represented", since such effects exist only at the sub-triple level of nodes and relations \cite{popularity-agnostic-eval}.

As such, they define a stratification procedure that first calculates link prediction performance in the context of all relationships individually \cite{popularity-agnostic-eval}. They then re-balance all of these results based on subject and object degree, and finally combine all relationship-specific performance metrics into a single performance score by re-weighting based on relationship frequency \cite{popularity-agnostic-eval}. They allow the degree of re-weighting to be configurable, meaning that they can choose to fully re-balance (i.e. removing all frequency bias in the test set), to counter-balance (inserting a bias inversely proportional to node / relation frequency) or not re-balance at all. They perform this operation on both the Hits@K and MRR metrics, resulting in new metrics called ``strat-Hits@k" and ``strat-MRR" \cite{popularity-agnostic-eval}.

Mohamed et al. (2020) then use their stratified link prediction metrics to evaluate 4 KGEMs (TransE, DistMult, ComplEx, and HolE) trained on FB15k and YAGO3-10 \cite{popularity-agnostic-eval}. Their results indicate that re-balancing to remove degree-related and relation-related biases results in a drop in reported KGEM performance \cite{popularity-agnostic-eval}. In other words, their re-evaluation exposed that low-degree nodes and low-frequency relations are learned substantially less reliably than those with higher degree / frequency \cite{popularity-agnostic-eval}. 

While they call out node-relationship and node-node co-frequencies as being relevant structural qualities, they do not specifically re-balance for these or empirically test for their impact on link prediction performance \cite{popularity-agnostic-eval}. Regardless, their results indicate very strongly that frequency-related structural metrics are of high relevance to the characterisation of KGs and, particularly, to the link prediction task.

\subsubsection{Kotnis et al. (2017)}
Kotnis et al. (2017) primarily focuses on the impact of negative sampler choice on link prediction using KGEMs, rather than on characterising link prediction in terms of KG structure \cite{neg-sampler-analysis}. Notwithstanding, in their analysis of which negative samplers work for different KGEMs and KGs, they find several important results regarding what elements of KG structure directly impact how well various negative sampling protocols work and, therefore, how well various KGEMs can learn \cite{neg-sampler-analysis}.

They perform an analysis of 6 negative sampling protocols on 4 KGEMs (TransE, DistMult, ComplEx, and RESCAL) trained and evaluated on 2 different KGs (FB15k-237 and WN18RR) \cite{neg-sampler-analysis}. Their results show that how effective various negative samplers are for the purpose of training KGEMs is based on \cite{neg-sampler-analysis}: 

\begin{itemize}
    \item \textbf{Relationship Frequency} -- They show that learning low-frequency relations is much less reliable than learning high-frequency relations, which leads to lower link prediction performance on low-frequency relations \cite{neg-sampler-analysis}. They further show that this effect persists generally regardless of the negative sampling strategy used \cite{neg-sampler-analysis}.
    \item \textbf{Node-Relationship Co-Frequency} -- They show that using pseudo-typed negative sampling is sensitive to node-relationship co-frequency. As the co-frequency decreases, the number of possible pseudo-typed corruptions necessarily decreases as well -- meaning that fewer negatives can be generated \cite{neg-sampler-analysis}. They show that this lack of negatives can lead to reduced LP performance \cite{neg-sampler-analysis}.
\end{itemize}

They further highlight that node degree and relationship frequency are critical to characterising KGs, and that node degree is expected to be partially determinant of link prediction from a theoretical perspective \cite{neg-sampler-analysis}.

\subsubsection{Rossi et al. (2021)}
Rossi et al. (2021) presents a general comparative overview of KGEMs for the link prediction task \cite{kges-for-lp-compare}. Specifically, they evaluate 16 different KGEMs (including TransE, DistMult, and ComplEx) on 5 different benchmark KGs (FB15k, FB15k-237, WN18, WN18RR, and YAGO3-10) \cite{kges-for-lp-compare}. They then define a few main structural and non-structural features and explore how each feature correlates to the ranks assigned to link prediction queries \cite{kges-for-lp-compare}. The 2 KG structural features they chose are:

\begin{itemize}
    \item \textbf{Number of peers}. Number of peers is what this paper refers to as ``node-relation co-frequency", except that it is defined in terms of how frequently a node and relation connect to a \textit{single} given other node \cite{kges-for-lp-compare}.
    \item \textbf{Relation path support} -- an estimate of how various paths (multi-hop) from the subject to the object in a triple contribute to its information content \cite{kges-for-lp-compare}.
\end{itemize}

Overall, their results indicate that triples with more possible alternatives for the node being predicted result in lower performance, as KGEMs struggle to distinguish between a larger set of possible nodes. The opposite effect also holds -- when there are many triples connecting to the same object (or subject), and that object (or subject) is being predicted, it is generally predicted with much higher accuracy. Finally, they show that higher relation path support of a triple leads to better predictions in almost all cases.

While they provide an analysis of their other KG features, we omit that analysis from this article as their other features are non-structural. For information on those, the reader is directed to their article \cite{kges-for-lp-compare}.

\subsubsection{Sardina et al. (2022)}
In the author's previous work, we analysed KGEM performance on link prediction as a function of graph structure \cite{centrality-and-hyperparams}. Specifically, we looked at 2 different ways of measuring structure, both of which are defined at the global (whole-KG) level:

\begin{itemize}
    \item \textbf{Distribution of Degrees:} We analysed the distribution of degrees in each KG studied in terms of its percentiles and the ratio of the highest-degree node to the number of triples \cite{centrality-and-hyperparams}.
    \item \textbf{Frequency of Nodes as Sources, Sinks or Repeats:} We defined source nodes as nodes that are only ever used as subjects, sink nodes as nodes that are only ever used as objects, and repeats as all other nodes (observed as a subject and object at least once) \cite{centrality-and-hyperparams}. We used the raw count of nodes in each group, as well as the ratio of those raw counts to the total number of triples in the graph, as characterisations of graph structure \cite{centrality-and-hyperparams}.
\end{itemize}

We then examined the performance of various KGEMs on a total of 9 biological KGs (all obtained from the Bio2RDF mashup of biological KGs), using sequential optimisation to select hyperparameter values to use when learning the KGs \cite{centrality-and-hyperparams}. We then built a Lasso regression model, using two types of structural features individually to correlate KG structure to KGEM performance. It is important to note, however, that the Lasso model was trained on \textit{all} of the data; in other words, we did not train-test split the Lasso model, but rather used it directly to ask whether finding correlation between structure and KGEM performance was possible.

Our results indicated we could predict KGEM performance given global KG structure in all cases with generally high, but variable, accuracy \cite{centrality-and-hyperparams}. In all cases, this performance was driven by a very small number (only 2 or 3) structural features. Finally, by analysing which features out Lasso model used to make predictions, we were able to provide evidence for a direct link between their chosen structural features and link prediction performance \cite{centrality-and-hyperparams}. In particular, the distribution of node degrees, the ration of sink nodes to triple, and the ration of repeat nodes to triples, were the three most important features they found to relate to KGEM performance \cite{centrality-and-hyperparams}.

\subsubsection{Sardina et al. (2024)}
In other previous work by the author, we created a model called TWIG to simulate the output of the KGEM ComplEx on the UMLS dataset in terms of KG structure and KGEM hyperparameters \cite{twig}. We first ran ComplEx 1215 different hyperparameter combinations (including 3 negative samplers and 3 loss functions) on the UMLS dataset. We then extract a total of 22 structural features (based on node degree, predicate frequency, node-relationship co-frequency, and node-node co-frequency) from a triple and all of its neighbouring triples \cite{twig}. We then combine both of these data streams and use them to predict KGEM performance both at the local level (how well a single triple can be learned) and at the global level (the overall performance of ComplEx on UMLS for a given set of hyperparameters) \cite{twig}.

Our results show that the results of ComplEx and UMLS can be accurately predicted at the level of global KGEM performance using the structural features we selected. However, we did not perform an ablation study to explore which structural features are important to KGEM performance \cite{twig}, which means we cannot yet make claims about which (if any) structural features were relevant to KGEM performance.

\subsubsection{Dave et al. (2024)}
Dave et al. (2024) explore how iteratively adding ontology-derived relations into the KG FB15k-237 affect how well it can be learned by different KGEMs \cite{struct-impact-on-kges}. This study is particularly unique in that it uses a \textit{structure-controlled} protocol, since iterative adding sets of edges changes KG structure in a controlled manner \cite{struct-impact-on-kges}. They then explored how well 6 different KGEMs (ComplEx, TransE, DistMult, RotatE, RESCAL, and TransR) were able to learn each structure-controlled variant of FB15k-237 \cite{struct-impact-on-kges}.

Their results show that adding in extra ontology-based relations to FB15k-237 generally results in decreased performance of the KGEMs tested \cite{struct-impact-on-kges}. While they do not provide a detailed numerical description of how adding in various relations affects the distribution of node degrees \cite{struct-impact-on-kges}, adding relations can only result in \textit{increasing} the degree of at least some nodes. As such, their results suggest that increasing degree of some nodes can have detrimental affects on learning -- something that echoes the results of Bonner et al. \cite{topological-imbalance}.

That said, the study has some limitations. In particular, they did not perform a hyperparameter search, and instead used a constant set of hyperparameters for all KGEMs and KG-structure variants tested \cite{struct-impact-on-kges}. In light of other works noting that hyperparameters are KG and KGEM dependent \cite{old-dog-new-tricks,dark-into-light}, this could have resulted in biased results of the relative performance on each model on each dataset.

\section{Hyperparameter Preference and Link Prediction Performance} \label{sec-hyp-performance-analysis}
This section provides a general overview of how hyperparameter choice has been shown to affect KGEM performance on LP in the literature. Table \ref{tab-hyp-studies} presents an overview of state-of-the-art literature that has explored hyperparameter preference in KGEMs. A detailed description of each of the studies, and their key findings, follows.

\begin{table*}[!ht]
    \centering
    \begin{tabular}{|l|p{1.8cm}|p{1.8cm}|p{1.3cm}|p{2.3cm}|p{2cm}|}
    \hline
        \textbf{Article} &\textbf{Scoring Functions} & \textbf{Negative Samplers} & \textbf{Losses} & \textbf{Optimisers} & \textbf{Others} \\ \hline
        Kotnis et al. (2017) \cite{neg-sampler-analysis} & ** & ** & . & . & . \\ \hline
        Sameh et al. (2019) \cite{loss-func-analysis} & ** & . & ** & . & . \\ \hline
        Jain et al. (2017) \cite{baselines-kges} & ** & . & . & . & ** \\ \hline
        Kadlec et al. (2020) \cite{baselines-kges-2} & ** & . & . & . & ** \\ \hline
        Ali et al. (2022) \cite{dark-into-light} & ** & . & ** & ** & * \\ \hline
        Ruffinelli et al. (2020) \cite{old-dog-new-tricks} & ** & * & ** & ** & ** \\ \hline
        Sardina et al. (2024) \cite{twig} & *** & **** & *** & *** & *** \\ \hline
    \end{tabular}
    \caption{An overview of studies in the literature documenting the effects of various hyperparameters on KGEM performance. ** = the given hyperparameter was directly evaluated in the context of link prediction; * = limited evidence for the hyperparameter's impact in LP was given; . = the given hyperparameter was not the subject of evaluation in that study.}
  \label{tab-hyp-studies}
\end{table*}

\subsubsection{Kotnis et al. (2017)}
Kotnis et al. (2017) provides an in-depth study of negative sampler preference in KGEMs -- specifically, how choice of negative sampler interacts with KGEM scoring functions and KG structure \cite{neg-sampler-analysis}. They examined 4 different KGEM scoring functions (TransE, DistMult, ComplEx, and RESCAL) paired with 6 negative samplers on 2 KGs (FB15k and WN18) \cite{neg-sampler-analysis}. They vary the number of negatives per positive on a grid of values, and report results for all negatives-per-positive values tested \cite{neg-sampler-analysis}. All other hyperparameters were determined either arbitrarily by the authors, or through a random hyperparameter search \cite{neg-sampler-analysis}. The negative sampling protocols they investigated are:

\begin{itemize}
    \item \textbf{Full Random} -- randomly replace the subject or object of a triple with any other node,
    \item \textbf{Corrupting Positive Instances} -- randomly replace the subject or object of a triple with any other node that has been observed as the subject of object of the triple's predicate before.
    \item \textbf{Typed Sampling} -- randomly corrupt the subject or object of a given training triple subject to type constraints to restrict what corruptions are valid,
    \item \textbf{Relational Sampling} -- randomly corrupting the predicate, rather than the subject or object,
    \item \textbf{Nearest Neighbour Sampling} -- using a pre-trained KGEM to suggest corruptions that are close to the correct answer in embedding space, and
    \item \textbf{Near Miss Sampling} -- using a pre-trained KGEM to create negatives that it estimates to be the hardest to learn.
\end{itemize}

Their experiments resulted in several conclusions about the nature of negative samplers in KGEMs:
\begin{itemize}
    \item \textbf{Preference for sampling more negatives.} They found that increasing the number of negatives generated per true triple almost always increases KGEM performance, regardless of the negative sampler, scoring function, or KG being used \cite{neg-sampler-analysis}. However, after a certain point, further increases to the number of negatives generated has little to no impact on performance \cite{neg-sampler-analysis}.
    \item \textbf{Dependency on KG connectivity.} They find that the primary determinant of negative sampler preference is how densely connected the graph is \cite{neg-sampler-analysis}. Node degree and node-relation co-frequency seem to be the most relevant measures of connectivity in this case \cite{neg-sampler-analysis}. They specifically highlight that low node-relation co-frequency in FB15k results in poor performance of pseudo-typed sampling, as low node-relation co-frequency means that there are not enough options for negative generation, which requires (lower-quality) pure random negatives to be generated instead \cite{neg-sampler-analysis}. On WN18, where this is not an issue, pseudo-typed sampling is typically optimal \cite{neg-sampler-analysis}.
    \item \textbf{Minor dependency on the scoring function.} The optimal negative sampler depends weakly on what KGEM scoring function is being used \cite{neg-sampler-analysis}. They find that more realistic negatives (i.e. those that are not pure random) tend to be better for most scoring functions, but that TransE, being so simplistic, does better on full random negative sampling \cite{neg-sampler-analysis}.
    \item \textbf{Minor dependency on relationship frequency.} They find that the negative sampling strategy that is globally optimal for a KG is not always the one that learns low-frequency relations the best, and that switching to a different negative sampler can result in learning lower-frequency relations more reliably. However, this effect is inconsistent, and generally the globally-optimal negative sampler for a given KG-KGEM combination is is also optimal for most low-frequency relations \cite{neg-sampler-analysis}.
\end{itemize}

\subsubsection{Sameh et al. (2019)}
Sameh et al. (2019) provide an in-depth analysis of loss functions used in KGEMs what factors determine loss function preference \cite{loss-func-analysis}. They examine how three KGEMs (TransE, DistMult, and ComplEx) and 5 KGs (FB15k-237, NELL50k, NELL50k239, WN18, and WN18RR) interact with 8 choices of loss functions. They categorise these loss functions into three categories:

\begin{itemize}
    \item \textbf{Pointwise losses:} In this category they consider Pointwise Hing Loss, Pointwise Logistic Loss, Pointwise Square Error Loss, and Pointwise Square Loss \cite{loss-func-analysis}.
    \item \textbf{Pointwise losses:} In this category they consider Pairwise Hinge Loss and Pairwise Logistic Loss \cite{loss-func-analysis}.
    \item \textbf{Multi-class losses:} They con Binary Cross Entropy Loss and Negative Log Softmax Loss \cite{loss-func-analysis}.
\end{itemize}

They train every combination of scoring function, loss function, and KG on their optimal hyperparameters (determined with a grid search) \cite{loss-func-analysis}. Following this, they analyse how each of their loss function (on its own and as a part of its category) influences KGEM performance on link prediction. The main results that they report are as follows:

\begin{itemize}
    \item \textbf{Loss function preference depends on the KGEM}. They show that TransE (as an additive model) does better when trained with pairwise losses than with pointwise losses \cite{loss-func-analysis}. For DistMult and ComplEx (multiplicative models), this trend is reversed -- they do better when trained with pointwise losses compared to pairwise losses \cite{loss-func-analysis}.
    \item \textbf{Some loss functions dominate others.} Some loss functions were found to be universally better than others. For example, Negative Log Softmax Loss was found to outperform Binary Cross Entropy Loss in all cases, and in most cases Pointwise Square Loss outperformed all other pointwise losses \cite{loss-func-analysis}.
\end{itemize}

\textbf{Kadlec et al. (2020) and Jain et al. (2017)}
Kadlec et al. (2020) and Jain et al. (2017) both perform very similar studies on KGEM benchmarking. They take various KGEMs and show that when re-training the common baseline KGEMs (such as DistMult) on better hyperparameter combinations, that they can perform the best reported results of more recent KGEMs in the literature \cite{baselines-kges,baselines-kges-2}. While the goal of their studies is not to compare different hyperparameter combinations, they do call out some trends in hyperparameter preference that became clear during their evaluation.

Specifically, Jain et al. (2017) and Kadlec et al. (2020) both highlight that increasing the number of negatives per positive used during negative sampling leads to better results in all cases \cite{baselines-kges,baselines-kges-2} -- therefore agreeing with the results presenting in Kotnis et al. (2017) \cite{neg-sampler-analysis}.

In addition to this, Kadlec et al. (2020) finds that increasing batch size always leads to better performance of KGEMs, and propose that higher batch size is generally ideal for achieving higher link prediction performance \cite{baselines-kges}.

\subsubsection{Ali et al. (2022)}
Ali et al. (2022) perform a large-scale evaluation of 21 KGEMs (including TransE, DistMult, and ComplEx) on 4 KGs (FB15k-237, WN18RR, Kinships, and YAGO3-10), running a large hyperparameter search on each KG-KGEM combination to determine optimal performance in all cases \cite{dark-into-light}. They then examine the overall performance of each KGEM and use their array of results to analyse relative efficacy of different KGEMs and hyperparameters \cite{dark-into-light}. Finally, they examine how well various logical relations in a KG (such as symmetry, transitivity, etc) can be modelled by each KGEM tested \cite{dark-into-light}. However, they do not analyse performance in terms of frequency-based structural metrics of KGs \cite{dark-into-light}. Overall, the results they obtain on hyperparameter preference and link prediction performance are as follows:

\begin{itemize}
    \item \textbf{Hyperparameters are dependent both on the KG and KGEM:} The optimal hyperparameters they found for each KG-KGEM pair indicate that optimal hyperparameters are a function of the the KG being used and the KGEM being used \cite{dark-into-light}. As such, knowing only the KG being learned, or only the KGEM being used, is not enough to determine optimal hyperparameters.
    \item \textbf{KGEM scoring functions strongly influence performance.} They provide evidence that the choice of scoring function has a huge impact on link prediction performance \cite{dark-into-light}. While they show that many models (when trained on their optimal hyperparameters) achieve similar results across different KGs, others are more variable and generally less-well performing \cite{dark-into-light}. Notably, they show that ComplEx and DistMult generally do quite well, and that TransE typically lags behind them \cite{dark-into-light}.
    \item \textbf{Different KGEMs are differently sensitive to different hyperparameter configurations.} They show that different KGEMs are more or less sensitive to the hyperparameter configuration chosen \cite{dark-into-light}. For example, ComplEx tends to achieve high performance on a wide range of hyperparameter combinations on all KGs \cite{dark-into-light}. DistMult and TransE trained on Kinships is highly sensitive to hyperparameter choice, but when trained on WN18RR and FB15k-237 their performance is less sensitive to hyperparameter choice \cite{dark-into-light}. Similar trends hold for other KGs / KGEMs tested \cite{dark-into-light}.
    \item \textbf{Loss function preference depends on the KGEM and KG.} They show that different loss functions result in better or worse link prediction performance depending both on the KGEM scouring functions and the KG being learned \cite{dark-into-light}. However, they do not provide a detailed analysis of this, or explore what elements of KG structure might be responsible for loss function preference being dependent on the KG being learned \cite{dark-into-light}.
    \item \textbf{The Adam optimiser generally outperforms Adadelta.} They find that the Adam optimiser tends to outperform Adadelta on the Kinships dataset across all KGEMs and hyperparameter combinations tested, and posit that this holds across other KGs as well \cite{dark-into-light}. 
\end{itemize}

\subsubsection{Ruffinelli et al. (2020)}
Ruffinelli et al. (2020) performs a large-scale evaluation of 7 KGEMs (including TransE, DistMult, and ComplEx) on 2 KGs (FB15k-237 and WN18RR) using a Bayesian (weighted) hyperparameter search for each KG-KGEM combination \cite{old-dog-new-tricks}. In doing so, they show that common KGEM baselines (such as DistMult and ComplEx) can outperform more recent KGEMs when trained on a better set of hyperparameters \cite{old-dog-new-tricks}. This leads them to question if the state-of-the-art has been producing better models, or just using better hyperparameters for newer models \cite{old-dog-new-tricks}.

Their overall results on hyperparameter preference and link prediction performance are outlined below:

\begin{itemize}
    \item \textbf{Cross entropy loss dominates other loss functions:} They show that, across all KGEMs tested on both FB15k-237 and WN18RR, that cross entropy loss outperforms all other loss functions \cite{old-dog-new-tricks}. Note that this finding does not necessarily contradict Sameh et al. (2019) \cite{loss-func-analysis} as they do not examine cross entropy loss; however, it does contradict Ali et al. (2022) \cite{dark-into-light}, who's results indicate that loss function preference is more nuanced.
    \item \textbf{Hyperparameters are dependent both on the KG and KGEM:} The optimal hyperparameters they determined for every KG-KGEM pair was distinct from that of other KG-KGEM pairs in general \cite{old-dog-new-tricks}. However, they do not analyse what aspects of a KG (structural or otherwise) might lead to different KGs having different influences on hyperparameter preference \cite{old-dog-new-tricks}. They show this effect for almost all hyperparameters studied, including the optimiser, regulariser, and other hyperparameters \cite{old-dog-new-tricks}.
    \item \textbf{Some hyperparameters are more important than others.} For each hyperparameter value found for each KG-KGEM pair, they also report the best competing configuration that assigns a different value to that hyperparameter \cite{old-dog-new-tricks}. While this is an imperfect comparison, they note that it can serve as a general proxy for the importance of each hyperparameter \cite{old-dog-new-tricks}. By applying this, they show that some hyperparameters (such as loss function) are more influential on overall MRR than others (such as the embedding dimension) \cite{old-dog-new-tricks}. They finally show that the relative importance of hyperparameters does sometimes vary based on the KG and KGEM being used \cite{old-dog-new-tricks}.
    \item \textbf{Hyperparameter sensitivity depends on both KG and KGEM.} They show that how sensitive link prediction results are to hyperparameters (i.e. how much performance changes between different hyperparameter configurations) is dependent both on the KG and KGEM being used \cite{old-dog-new-tricks}. However, in general sensitivity to hyperparameter values is more influenced by the KGEM than by the KG it is trained on \cite{old-dog-new-tricks}.
    \item \textbf{KGEM scoring functions influence performance.} They show that the scoring function used is strongly influential on overall model performance \cite{old-dog-new-tricks}. Specifically, they highlight that many models achieve very similar results when trained on their optimal hyperparameters, but that some KGEMs (such as TransE) generally lag behind the others in performance \cite{old-dog-new-tricks}.
\end{itemize}

Despite the quite strong and far-reaching results of their study, two limitations remain. First off, their experiments on negative samplers did not include most common negative sampling strategies \cite{old-dog-new-tricks}. Instead, they compared full random negative sampling against all possible negatives \cite{old-dog-new-tricks} -- which means that their analysis of negative sampling preference cannot necessarily be expected to apply to how different negative sampling protocols interact with each other.

Second, their comparison of hyperparameter sensitivity is potentially biased. Since they use a weighted hyperparameter search for all values, rather than a full random or grid search, the hyperparameter values they sample are potentially biased \cite{old-dog-new-tricks}. This means that assessing their variance (to determine hyperparameter sensitivity) or treating them as ablations (to assess the relative important of each hyperparameter) is also potentially biased \cite{old-dog-new-tricks}. As such, while their results on hyperparameter sensitivity and importance are powerful, they must be checked by acknowledgement that they come from a weighted Bayesian optimisation protocol, rather than an unweighted (full random or grid search) protocol.

\subsubsection{Sardina et al. (2024)}
In the author's previous work, as outlined in the Section \ref{sec-kg-struct-measurse}, we used a model called TWIG to simulate KGEM output on a given KG based on knowledge of the KGEM's hyperparameters the KG's structure \cite{twig}. Specifically, we used ComplEx as a KGEM, and UMLS as a KG, but did not study any other KGEM-KG combinations \cite{twig}. We showed that using hyperparameter and structural features is sufficient to predict the overall performance of KGEMs with high accuracy \cite{twig}. We then used this to suggest that there is a deterministic relationship between KG structure, KGEM hyperparameters, and KGEM performance \cite{twig}.

However, despite showing that there is a clear relationship between KG structure, hyperparameters, and KGEM performance, we did not provide a characterisation of \textit{which} elements of structure or \textit{which} hyperparameters correlate to these effects, something that limits direct analysis of our findings in this paper \cite{twig}. This is understandable seeing as the purpose of our study was not to provide a map of such inter-relationships. Our primary goal was instead to shown that (at least for ComplEx and UMLS) it is possible to find a deterministic function relation structure and hyperparameters to KGEM performance \cite{twig}. As a result, while our work is notable for its structure- and hyperparameter- based approach, it does not result as yet in direct evidence for any specific hyperparameter influence on KGEM performance.

\section{Conclusion and Open Directions}
This paper presents, to the extent of the knowledge of the authors, the first general survey of KG structure, hyperparameter preference, and KGEM performance on the link prediction task. It provides a detailed overview of all structural and hyperparameter dependencies that have been annotated in the literature, and summarises the state-of-the-art experiments in this area. It is the hope of the authors that this survey be of use for the development of new KGEMs, for the deeper analysis of exploration of existing KGEMs, and for enhancing structural characterisation and analysis of KGs in general.

In light of the survey presented here, we to conclude with a short list of what we consider the most relevant open research directions in the domain of KG structure and link prediction.

First, existing research has shown many relationships between structure and LP performance, and between hyperparameters and LP performance. However, very few have examined how KG structure interacts with hyperparameter preference. We believe that further research in this area would be useful to further characterise KGEMs and link prediction.

Second, this work does not consider the \textit{ontological} properties of KGs, such as frequencies of transitive / symmetric / asymmetric relationships (and so on). Nor does it consider node / relationship typing or type hierarchies. As ontologies are core to KG data representation, an exploration of how ontological properties of a graph interact with KG structure, hyperparameter preference, and link prediction performance wold be of great contribution to the field.

Finally, structural analysis of KGs is limited by a lack of diverse, structurally-controlled benchmark KGs. While work in this area exists (see Dave et al. (2024) \cite{struct-impact-on-kges}), and while significant work has been put into structure-controlled KG generation (i.e. PyGraft \cite{pygraft}), there is no consensus in this area yet. We propose that the establishment of desiderata for such standard graphs, as well as their publication in a standard library (such as PyKEEN \cite{pykeen}), would be a massive contribution to the state-of-the-art in KGEMs and LP.

\section*{Acknowledgements}
This research was conducted with the financial support of Science Foundation Ireland D-REAL CRT under Grant Agreement No. 18/CRT6225 at the ADAPT SFI Research Centre at Trinity College Dublin, together with sponsorship of Sonas Innovation Ireland.  The ADAPT SFI Centre for Digital Content Technology is funded by Science Foundation Ireland through the SFI Research Centres Programme and is co-funded under the European Regional Development Fund (ERDF) through Grant \# 13/RC/2106\_P2.

\bibliographystyle{splncs04}
\bibliography{main}

\end{document}